\pdfoutput=1
\documentclass[10pt,twocolumn,letterpaper]{article}

\usepackage{cvpr}
\usepackage{times}
\usepackage{epsfig}

\usepackage{cite}
\usepackage{graphicx}
\usepackage{amsmath}
\usepackage{amssymb}  
\usepackage{algorithmicx}
\usepackage{algorithm}
\usepackage{algpseudocode}
\usepackage{mathtools}
\usepackage{url}
\usepackage[table]{xcolor} 
\usepackage{threeparttable}
\usepackage{caption}
\usepackage{subcaption}
\usepackage{makecell}
\usepackage{pifont}
\usepackage{multirow}
\usepackage{booktabs}
\usepackage{chngcntr}

\usepackage[pagebackref=true,breaklinks=true,letterpaper=true,colorlinks,bookmarks=false]{hyperref}
\hypersetup{
    colorlinks=true,
    linkcolor=blue,
    filecolor=magenta,      
    urlcolor=cyan,
}

\newcommand\blfootnote[1]{%
  \begingroup
  \renewcommand\thefootnote{}\footnote{#1}%
  \addtocounter{footnote}{-1}%
  \endgroup
}

\counterwithin{paragraph}{subsection}

\cvprfinalcopy 

\def\cvprPaperID{****} 

\ifcvprfinal\fi
\begin{document}

\title{HandyNet: A One-stop Solution to Detect, Segment, Localize \& Analyze \\Driver Hands}

\author{Akshay Rangesh \qquad Mohan M. Trivedi\\
Laboratory for Intelligent \& Safe Automobiles, UC San Diego\\
{\tt\small \{arangesh, mtrivedi\}@ucsd.edu}
}

\maketitle

\blfootnote{$^\dagger$\href{https://www.youtube.com/watch?v=4dxSdFbnTFM&list=PLUebh5NWCQUah_cBzcRlZvoSMa-7GC3FL}{Video results}}
\blfootnote{$^\ddagger$\href{https://drive.google.com/open?id=1wV8gmTgap24MTFxCqno4_TLiB-3YPcc-}{Dataset}}
\blfootnote{$^\ddagger$Code: \href{https://github.com/arangesh/HandyNet}{https://github.com/arangesh/HandyNet}}

\begin{abstract}
Tasks related to human hands have long been part of the computer vision community. Hands being the primary actuators for humans, convey a lot about activities and intents, in addition to being an alternative form of communication/interaction with other humans and machines. In this study, we focus on training a single feedforward convolutional neural network (CNN) capable of executing many hand related tasks that may be of use in autonomous and semi-autonomous vehicles of the future. The resulting network, which we refer to as HandyNet, is capable of detecting, segmenting and localizing (in 3D) driver hands inside a vehicle cabin. The network is additionally trained to identify handheld objects that the driver may be interacting with. To meet the data requirements to train such a network, we propose a method for cheap annotation based on chroma-keying, thereby bypassing weeks of human effort required to label such data. This process can generate thousands of labeled training samples in an efficient manner, and may be replicated in new environments with relative ease. 
\end{abstract}

\section{Introduction}

The past decade has seen a rapid increase and improvement in the automation offered on consumer automobiles. This may be attributed to a corresponding growth in the technology and engineering required to make such automation reliable enough for widespread deployment. Based on the definitions provided by SAE International, vehicles with \textit{partial} and \textit{conditional} automation are already used in notable numbers across the world. Such vehicles would only become more ubiquitous with the improvement in technology, increase in the number vehicles offered with some form of autonomy, and the dwindling costs associated with such vehicles. Moreover, vehicles with automation will see a continued growth in the level of automation offered, with the end goal being complete automation - a situation where the driver is just another passenger. However, until a point of complete automation is reached, the perils of partial automation need to be dealt with.

\begin{figure}[t]
\begin{center}
\includegraphics[width=0.7\linewidth]{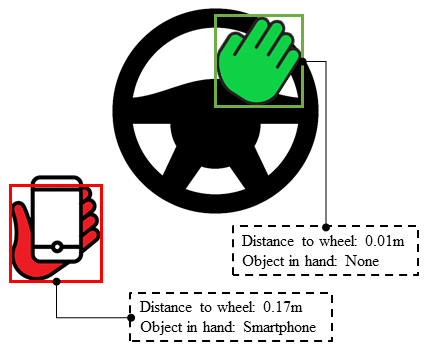}
\end{center}
\caption{Illustration of the intended goal our research. The proposed method is capable of detecting and segmenting driver hands, localizing them in 3D, and identifying the object each hand may be holding.}
\label{fig:introduction}
\end{figure}

The dangers of partial automation primarily arise from the need for the driver to be constantly monitoring the drive, or in the case of conditional automation, for the driver to be ready to take-over control at any given time. Yet decades of human factors research has shown that humans are not particularly good at tasks that require vigilance and sustained attention over long periods of time\cite{kyriakidis2017human}. It is also well known that rising levels of automation will lead to declining levels of awareness on the part of the human operator\cite{casner2016challenges}. This seems to suggest that it is not a matter of \textit{if}, but \textit{when} a driver will resort to non-ideal behavior, especially since most automated systems are designed to free the driver to do something else of interest. Thus, it is of extreme importance to monitor the driver and assess his/her readiness to take control in case of an unexpected failure of the system. This is the irony of automation, whereby the more advanced a system is, the more crucial may be the contribution of the human operator.

In this study, we propose a system that goes a long way towards monitoring a driver and assessing his/her readiness. In particular, we monitor and analyze driver hands beyond what is currently possible with just tactile sensors on the steering wheel. The main contributions of this paper are as follows: We propose a convolutional neural network capable of detecting and segmenting driver hands from depth images. This makes localizing the hands (in 3D) inside a vehicle cabin possible, which in turn enables the calculation of the distance of each hand to critical control elements like the steering wheel. In addition to this, the network is capable of identifying objects held by the driver during a naturalistic drive. All this is made feasible by a form of semi-automatic data labeling inspired by chroma-keying which we describe in detail.

\section{Related Work}
To the best of our knowledge, there is no existing work that addresses the problem of segmenting instances of driver hands in a naturalistic driving setting. We therefore provide a brief overview of works that pertain to driver and/or passenger hands inside a vehicle cabin.

The work in~\cite{das2015performance} outlines common challenges associated with detecting driver hands in RGB images in addition to proposing a dataset to train such detectors. The authors train and evaluate a few such detectors based on Aggregate Channel Features (ACF) as outlined in~\cite{dollar2014fast}. In addition to just detecting driver hands, the authors in~\cite{rangesh2016hidden} also track both hands of the driver using a tracking-by-detection approach. They also leverage common hand movement patterns to stably track hands through self occlusions. Moving over to the domain of human computer interaction, studies like \cite{molchanov2015hand, ohn2014hand, deo2016vehicle} identify driver hand gestures from a sequence of depth images. These methods directly produce the identified gesture from raw depth images and do not localize driver hands as an intermediate step. There have also been numerous contributions in the field of hand pose estimation, both inside a vehicle and otherwise. We refer the reader to~\cite{supanvcivcdepth} for a detailed survey on this subject. Finally, there has been noticeable work on analyzing driver hand activity. In~\cite{ohn2013vehicle}, the authors extract hand-designed features around pre-defined regions of interest like the steering wheel, infotainment control etc. to identify regions with high hand activity. The authors in~\cite{ohn2014beyond} take a different approach to identifying regions of activity by detecting and tracking hands for short periods of time and analyzing the temporal dynamics of hand locations. They also go on to detect abnormal events and activities. More recently, the authors in~\cite{borghi2018hands} have proposed a unique dataset for detecting and tracking driver hands inside vehicles. This dataset is captured using a Leap Motion device mounted behind the steering wheel. Although this approach makes detecting hands on the wheel much simpler, it also forgoes the capability to know where the driver hands are, if not on the wheel. 

In all studies listed above, the major limitation arises from the inherent depth ambiguity. The input to most methods are RGB images, severely restricting the utility of the outputs they produce. Simply localizing hands in 2D does not inform us about crucial information like the 3D distance to different control elements inside a vehicle cabin. Moreover, all these methods are heavily dependent on the camera view used for training, and re-training for new views would require devoting considerable efforts towards ground truth annotation. Even for methods relying on depth data, the end goal is achieved without actually localizing the hands. This is primarily due to the lack of sufficiently labeled depth data to train such algorithms. In this study, we overcome all these stumbling blocks and produce thousands of labeled depth images with relative ease. This method may also be replicated with little effort in new environments, and for different camera views.

\begin{algorithm}
\caption{Pseudocode for obtaining instance masks for a given sequence by chroma-keying}\label{alg:instances}
\textbf{Input:} $\{rgb_i^{reg}$, $depth_i^{reg}\}_{i=1}^{N}$ 

\Comment{registered RGB and depth for each frame $i$}

\textbf{Output:} $\{inst\_masks_i\}_{i=1}^N$ 

\Comment{binary mask for each hand instance in every frame}

\begin{algorithmic}[h]
\Statex
\For{$i \gets 1$ \textbf{to} $N$}
\State $r_i^{reg} \gets rgb_i^{reg}(:, :, 1)$ \Comment{red channel}
\State $g_i^{reg} \gets rgb_i^{reg}(:, :, 2)$ \Comment{green channel}
\State $b_i^{reg} \gets rgb_i^{reg}(:, :, 3)$ \Comment{blue channel}
\State $Y_i^{reg} \gets 0.3 \cdot r_i^{reg}+0.59 \cdot g_i^{reg}+0.11 \cdot b_i^{reg}$ 
\Statex \Comment{relative luminance}
\State $masks \gets (g_i^{reg} - Y_i^{reg}) \geq threshold$
\State $\{mask_j\}_j \gets CCL(masks)$
\Statex \Comment{connected component labeling}

\Statex
\Statex /*The block below is for cases where two instances might be merged in 2D, but are disjoint in 3D*/
\For{\textbf{each }$mask_k \in\{mask_{j}\}_{j}$}
\State $depth\_pixels \gets depth_i^{reg}(mask_k)$
\Statex \Comment{depth pixels corresponding to each mask}
\State $opt\_thresh \gets otsu(depth\_pixels)$ 
\Statex \Comment{get optimal threshold using Otsu's method} 
\State $mask_{k_1} \gets (depth\_pixels \geq opt\_thresh)$
\State $mask_{k_2} \gets (depth\_pixels < opt\_thresh)$
\State $\{mask_{j}\}_{j} \gets \{mask_{j}\}_{j}$\textbackslash$\{mask_k\}$
\State $\{mask_{j}\}_{j} \gets \{mask_{j}\}_{j} \bigcup \{mask_{k_1}, mask_{k_2}\}$
\State 
\EndFor

\Statex
\Statex /*The block below is for cases where a hand might be partially occluded resulting in disjoint regions*/
\Statex \textbf{Require: }{For each element in $\{mask_j\}_j$, the area in pixels is known.}
\State $inst\_masks_i \gets \{\}$
\While{$\{mask_j\}_j \neq \phi$}
\State $mask_{j'} \gets$ largest $mask \in \{mask_j\}_j$
\If{$Area(mask_{j'}) \leq 20$}
\State $\{mask_{j}\}_{j} \gets \{mask_{j}\}_{j}$\textbackslash$\{mask_{j'}\}$
\State \textbf{continue}
\EndIf
\For{\textbf{each }$mask_k \in\{mask_{j}\}_{j}, k \neq j'$}
\State $dist \gets Distance\_3d(mask_k, mask_{j'})$
\Statex \Comment{distance is calculated between centroids using eq.~\ref{eq:1}}
\If{$dist \leq 7cm$}
\State $\{mask_{j}\}_{j} \gets \{mask_{j}\}_{j}$\textbackslash$\{mask_k\}$
\State $\{mask_{j}\}_{j} \gets \{mask_{j}\}_{j}$\textbackslash$\{mask_{j'}\}$
\State $mask_{j'} \gets mask_{j'} \bigcup mask_k$
\Statex \Comment{combine $mask_k$ and $mask_{j'}$}
\EndIf
\EndFor
\State $inst\_masks_i \gets inst\_masks_i \bigcup \{mask_{j'}\}$
\EndWhile
\EndFor
\end{algorithmic}
\end{algorithm}

\section{Semi-Automatic Labeling based on Chroma-Keying}\label{sec:chromakey}

The key contribution of our work is a method for generating large amounts of labeled data in a relatively cheap manner for the task of hand instance segmentation. As is well known, deep learning methods although extremely powerful and accurate, require large amounts of labeled data to learn and generalize well. This requirement becomes even more unwieldy for tasks like semantic and instance segmentation, where pixel level annotations are required. Such tasks entail several hundred hours of human effort to generate enough samples for successfully training networks. These difficulties are usually overcome by hiring large groups of human ``annotators", either directly or through a marketplace such as the Amazon Mechanical Turk. This approach has its own limitations. First, this requires some form of monetary incentive which may be beyond the resources that are available. More importantly, networks trained on a particular dataset tend to perform best on similar data. Therefore, to ensure similar performance on the same task for a different set of data, more retraining on such data would be required; this leads to more expensive annotations. In this section, we describe a form of semi-automatic labeling based on \textit{chroma-keying}. This method can be replicated in different environments and even for different tasks with relative ease.

\subsection{Acquiring Instance Masks}
Chroma-keying is a technique popular in the visual effects community for layering images i.e. separating specific foregrounds from a common background based on color hues. This usually involves the use of green screens and body suits that visually contrast the object of interest from the other elements of scene, and hence can be separated with ease. 

\begin{figure}[t]
\begin{center}
\includegraphics[width=0.72\linewidth]{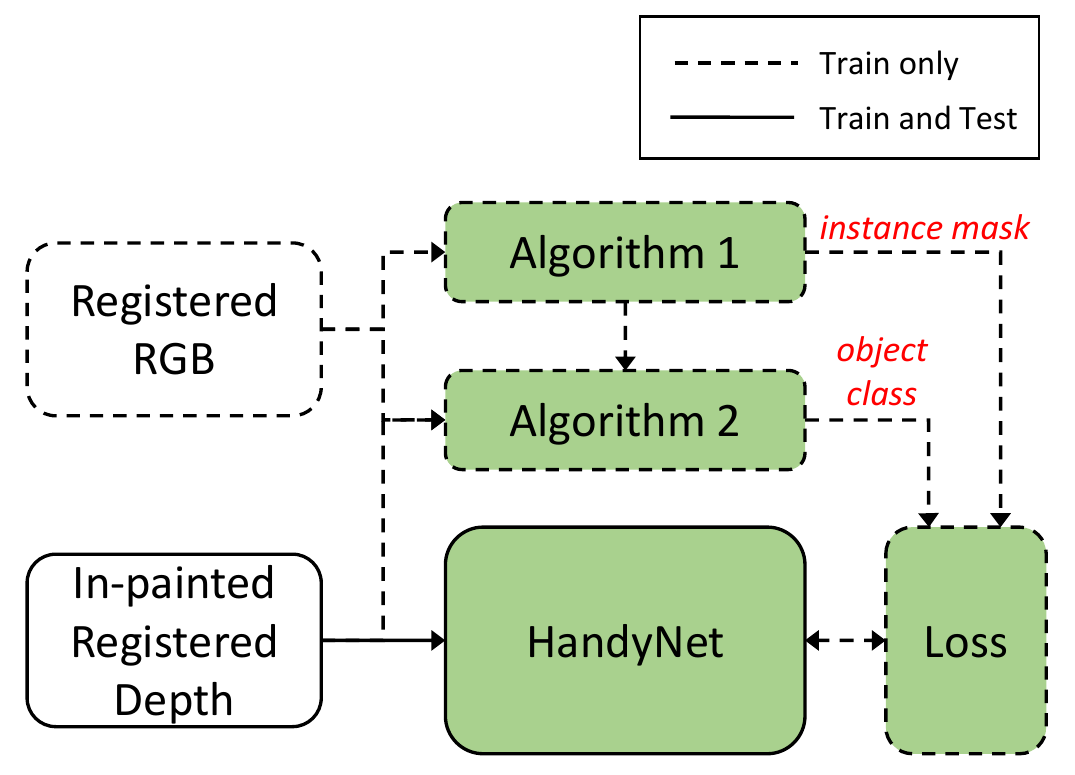}
\end{center}
\caption{Block diagram depicting proposed training and testing methodology.}
\label{fig:flow_diag}
\end{figure}

\begin{figure*}[!t]
\captionsetup[subfigure]{justification=centering}
  	\centering
  	\begin{subfigure}[t]{0.2\textwidth}
		\centering
		\includegraphics[width=\linewidth]{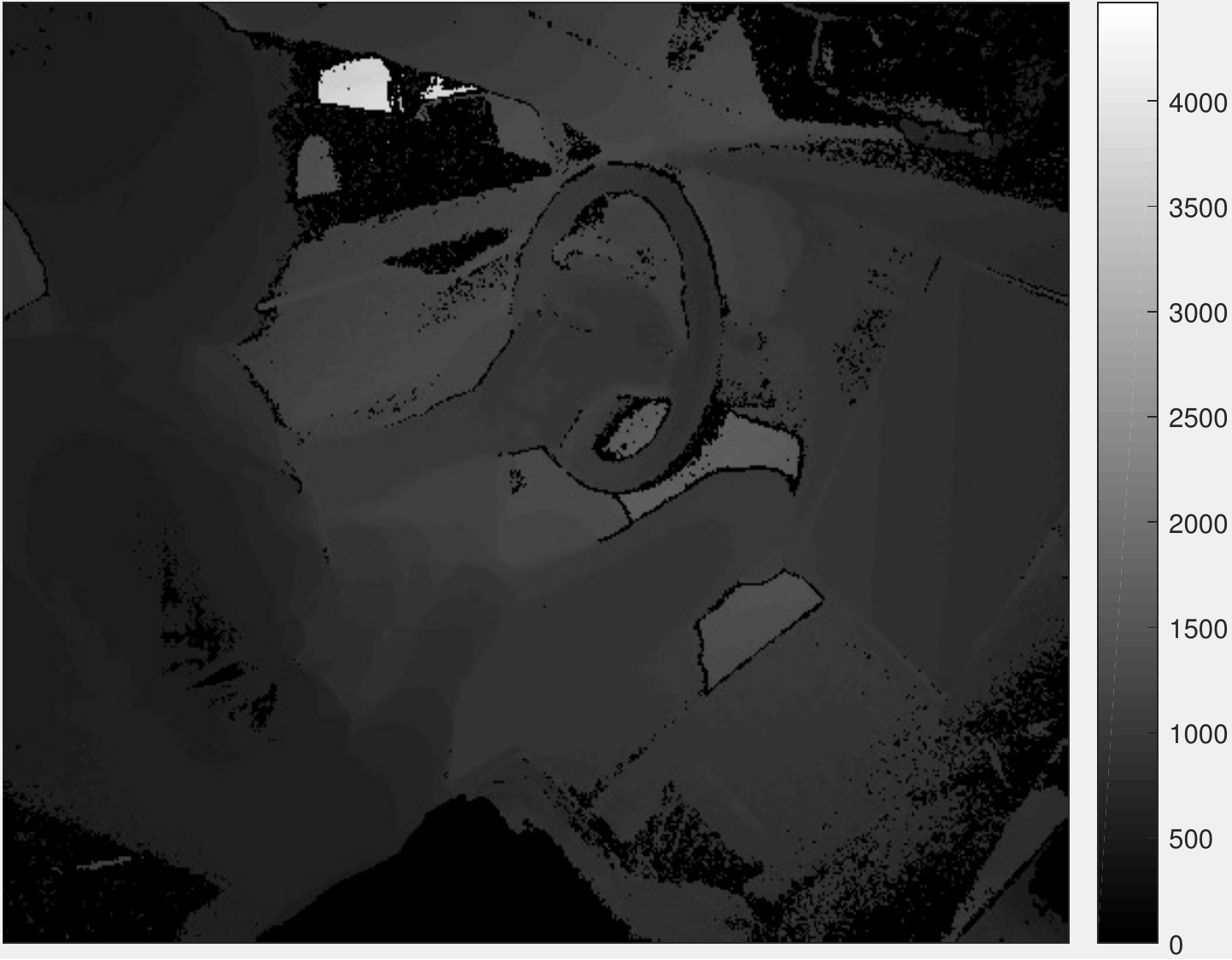}
		\caption{Undistorted, registered depth image}
	\end{subfigure}%
~  	
	\begin{subfigure}[t]{0.2\textwidth}
		\centering
		\includegraphics[width=\linewidth]{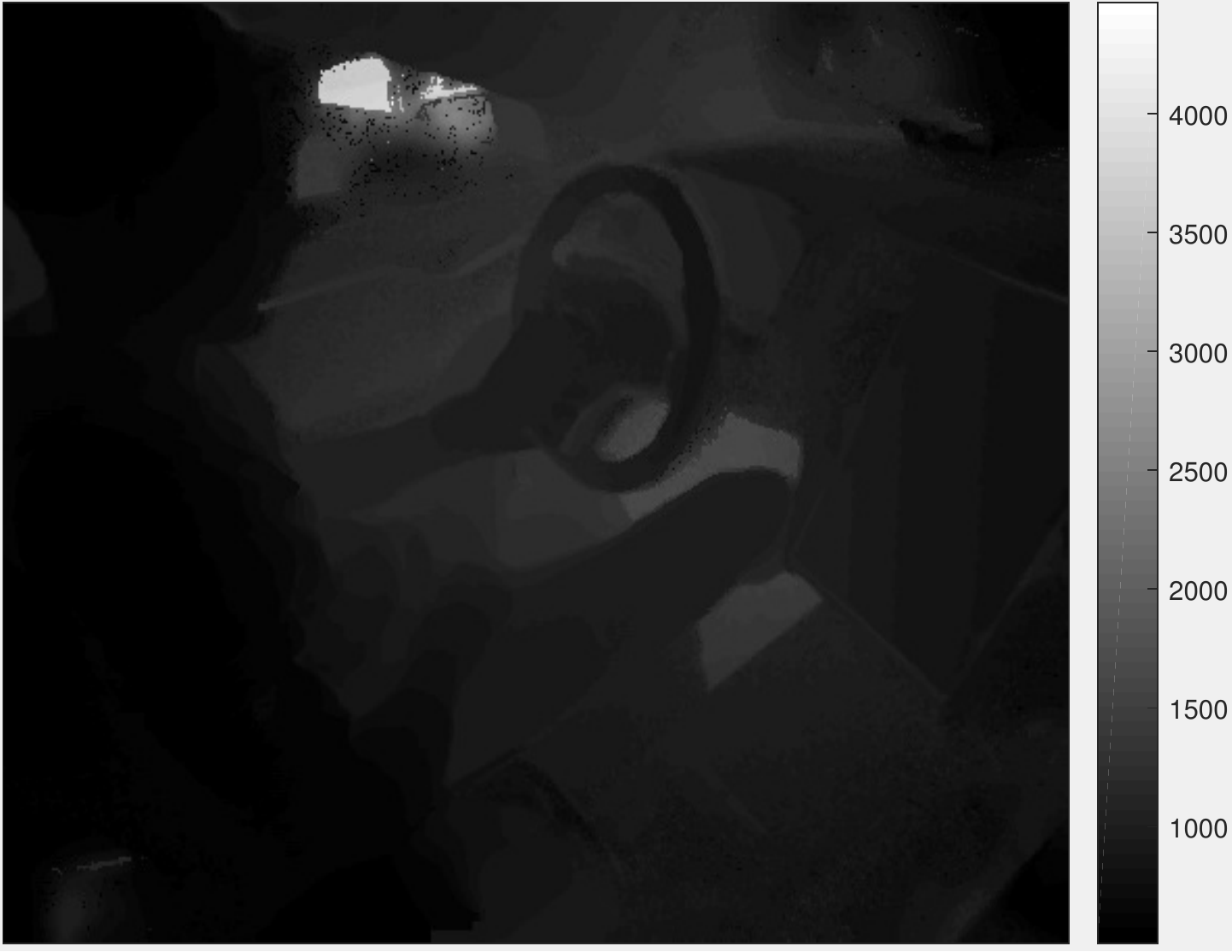}
		\caption{Undistorted, registered, in-painted depth image}
	\end{subfigure}
	\hspace{1mm}
~
	\begin{subfigure}[t]{0.18\textwidth}
		\centering
		\includegraphics[width=\linewidth]{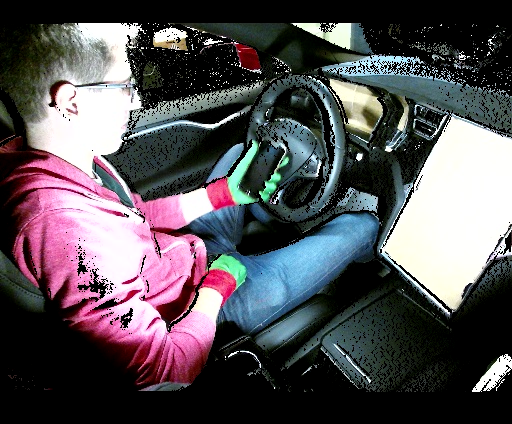}
		\caption{Registered RGB image}
	\end{subfigure}%
~	
	\begin{subfigure}[t]{0.18\textwidth}
		\centering
		\includegraphics[width=\linewidth]{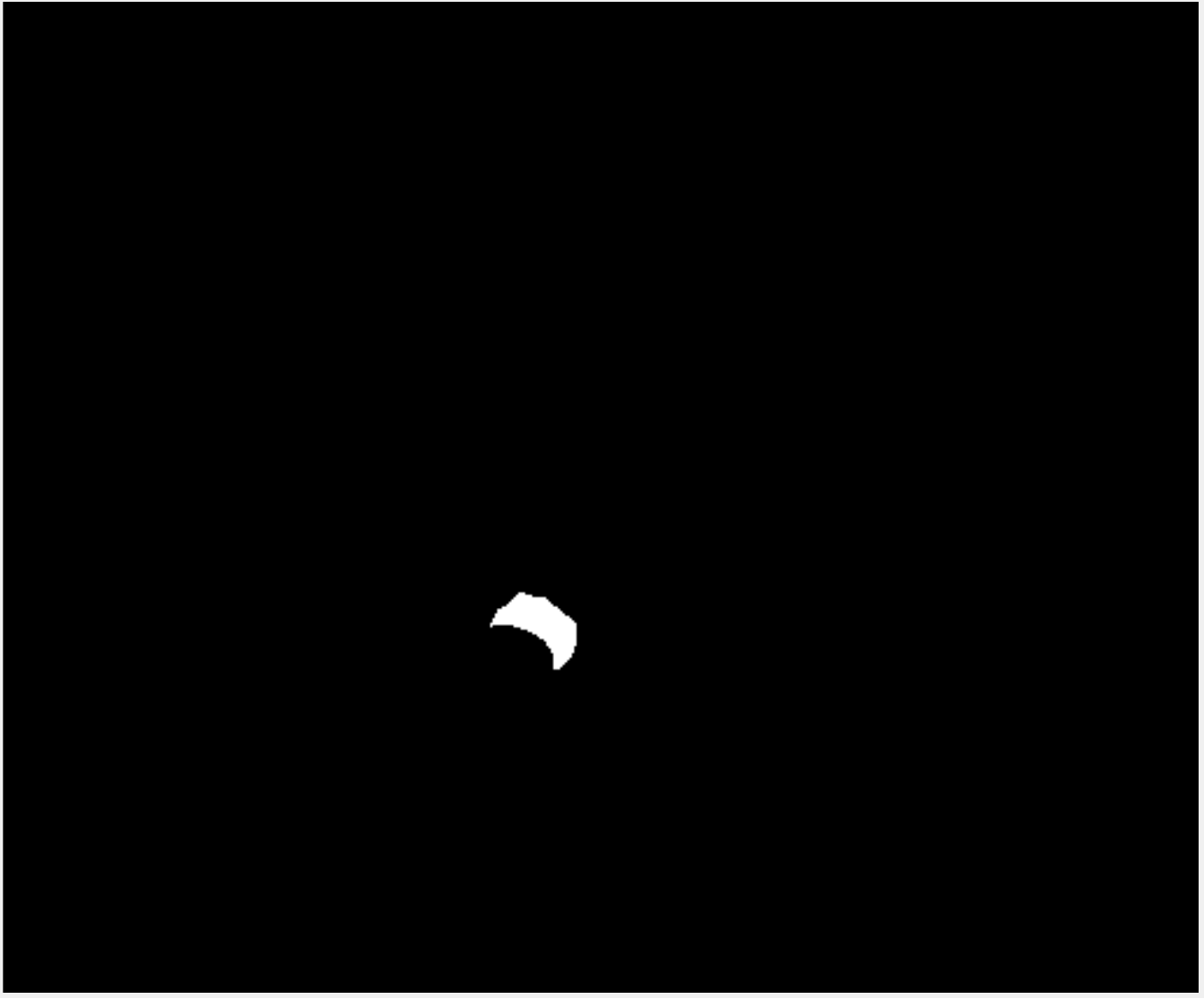}
		\caption{Hand instance with label \textit{no object}}
	\end{subfigure}%
~  	
	\begin{subfigure}[t]{0.18\textwidth}
		\centering
		\includegraphics[width=\linewidth]{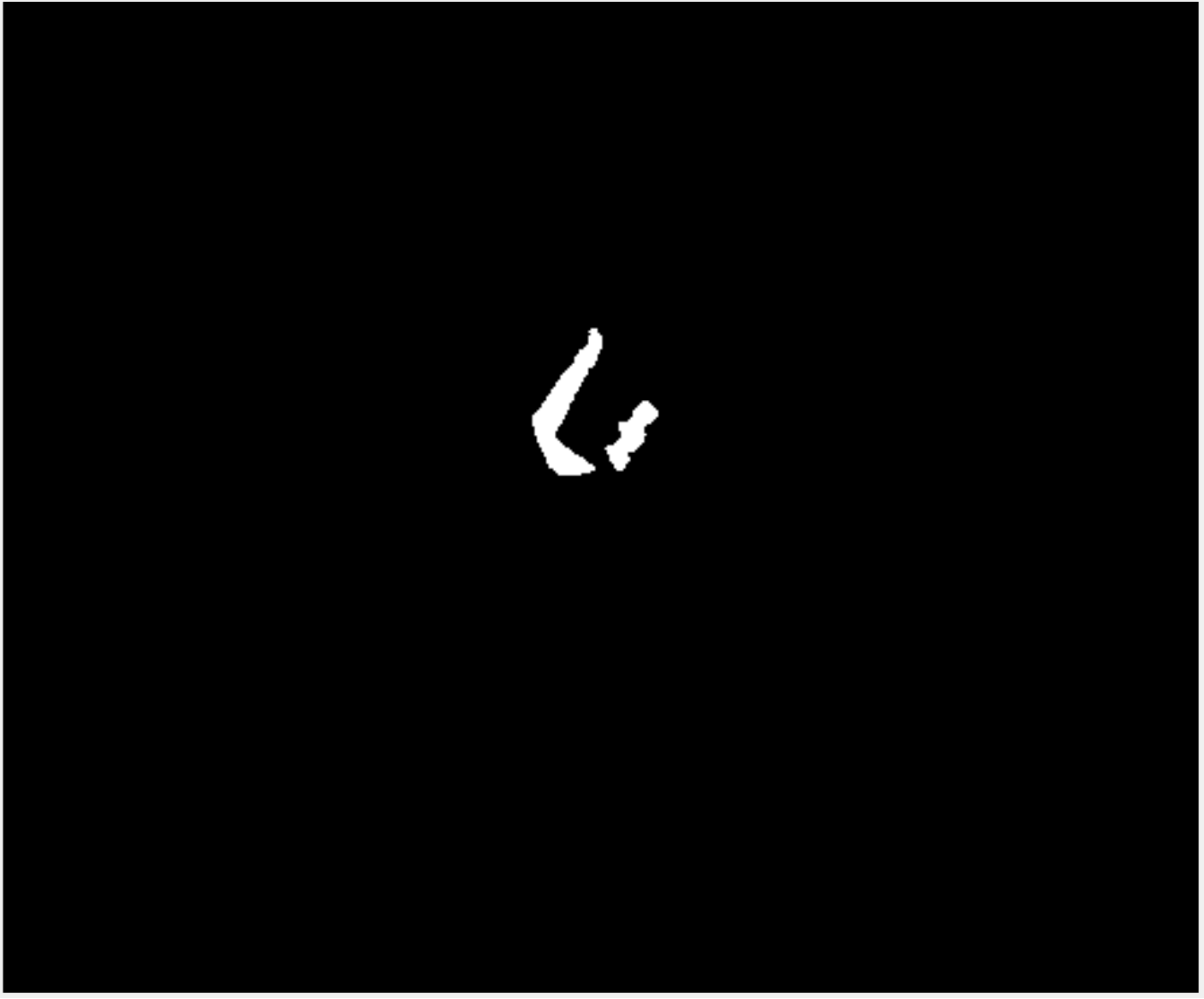}
		\caption{Hand instance with label \textit{smartphone}}
	\end{subfigure}
	\caption{\textbf{Input and desired outputs used to train the proposed network:} The input to the network (shown in (b)) is obtained after smoothing the raw undistorted depth (shown in (a)), while the desired outputs ((d) and (e)) are obtained from the registered RGB image (shown in (c)) using Algorithms~\ref{alg:instances} and \ref{alg:objects}.}
	\label{fig:input_output}
\end{figure*}

To leverage this technique for the task at hand, we make use of a Kinect v2 sensor comprising of an RGB and infrared (depth) camera with a small baseline. Note that we chose a Kinect for its ease of use, excellent community support and high-resolution depth images; however, any calibrated pair of RGB and depth cameras may be used. The input to the proposed network is the in-painted, registered version of the raw depth image, and the desired outputs are the instance masks for each hand of the driver. The registered RGB and depth images are captured using Kinect drivers provided by OpenKinect~\cite{wikiur}, and the depth images are in-painted by applying cross-bilateral filters at three image scales~\cite{silberman2012indoor}. The instance masks for supervision during training are obtained by chroma-keying the registered RGB images using the procedure detailed in Algorithm~\ref{alg:instances}. This requires the driver (subject) to wear green gloves as shown in Figure~\ref{fig:input_output}. We also make the subjects wear red wrist bands to ensure a clear demarcation for where the wrist ends and the hand begins. Additionally, we add soft lights inside the vehicle cabin to ensure the green gloves are uniformly lit. Note that good lighting is extremely important for accurate chroma-keying. Registering the RGB and depth images results in a one-to-one correspondence between every pixel in both images. This ensures that the masks obtained by chroma-keying the registered RGB images are valid supervision for the registered depth images. This way, the registered RGB images are only used for labeling hand instances during training, and are unused when the trained network is deployed. This entire procedure is illustrated in Figure~\ref{fig:flow_diag}, and example inputs and labels generated by this procedure are shown in Figure~\ref{fig:input_output}.

\begin{algorithm}
\caption{Pseudocode for labeling handheld objects for a given sequence}
\label{alg:objects}
\textbf{Require: }The driver holds the same object using the same hand for the entire duration of the sequence

/*Note that only one instance per frame is assigned $label$; other instances are assigned $0$ corresponding to \textit{no handheld object}. This is valid by the requirement stated above.*/

\textbf{Input:} $\{inst\_masks_i\}_{i=1}^N$, where

\hspace{2cm} $inst\_masks_i = \{mask_i^1, \cdots, mask_i^{M_i}\}$, 

\Comment{binary mask for each hand instance in every frame}

\hspace{1cm} $label \in\{1, 2, 3, 4\}$, 

\Comment{label for the object used in the sequence}

\hspace{1cm} $m_1 \in \{1, \cdots, M_1\}$, 

\Comment{user provided instance associated with $label$ for the first frame} 

\textbf{Output:} $\{o_i\}_{i=2}^N$

\Comment{object label for each hand instance in every frame}

\begin{algorithmic}[1]
\Statex
\State $o_1 \gets \{0\}^{M_1}$ \Comment{initialize all instances with zeros}
\State $o_1(m_1) \gets label$ 
\Statex \Comment{assign $label$ to instance holding object}
\State $last \gets mask_i^{m_1}$
\Statex \Comment{store last instance holding object}
\Statex /*Find instance in current frame closest to last known instance holding object*/
\For{$i \gets 2$ \textbf{to} $N$}
\State $min\_dist \gets \infty$
\For{$j \gets 1$ \textbf{to} $M_i$}
\State $cur\_dist \gets Distance\_3d(mask_i^j, last)$
\Statex \Comment{distance is calculated between centroids using eq.~\ref{eq:1}}
\If{$cur\_dist \leq min\_dist$}
\State $m_i \gets j$
\State $min\_dist \gets cur\_dist$
\EndIf
\EndFor
\State $o_i \gets \{0\}^{M_i}$ \Comment{initialize all instances with zeros}
\Statex /*The following condition handles cases where the hand holding the object is temporarily occluded*/
\If{$min\_dist \leq 15cm$}
\State $last \gets mask_i^{m_i}$
\State $o_i(m_i) \gets label$ 
\EndIf
\EndFor
\end{algorithmic}
\end{algorithm}

At this point, we would like to point out two caveats of this approach. First, two hand instances are considered unique when their distance in 3D is sufficiently large. This then implies that two instances very close to each other (for example, when a driver clasps both hands) cannot be separated. We do not try to separate such instances and force the proposed network to detect such \textit{merged} instances. Alternatively, one could make each subject wear two differently colored gloves. Second, one may argue that any algorithm trained on data with subjects wearing gloves and captured under controlled lighting may not generalize well to real world scenarios. We disprove such arguments by providing large amounts of qualitative results (images and videos) on multiple hours of real world drives, with subjects not present in the training split.

Although using depth images as input allows us to use chroma-keying for cheap supervision, we are motivated by other factors as well. First, using depth images as input circumvents any privacy issues that may arise with having cameras inside cars. Second, depth cameras are relatively unaffected by harsh illumination or lack thereof (e.g. during nighttime driving). Finally, once driver hands are located in depth images, it is straightforward to calculate where the hands are in 3D, and how far they are from critical control elements like the steering wheel. Such information may be extremely useful to gauge the readiness of a driver to \textit{takeover} control from a semi-autonomous/autonomous vehicle. 

For a pixel at $(x, y)^T$ in an undistorted depth image, the corresponding 3D point location $(X, Y, Z)^T$ is obtained as follows:
\begin{equation}\label{eq:1}
X = \frac{(x - c_x) \cdot d}{f_x},
Y = \frac{(y - c_y) \cdot d}{f_y},
Z = d,
\end{equation}
where $d$ is the depth value at pixel $(x, y)^T$, $(c_x, c_y)^T$ is the principal point, and $\{f_x, f_y\}$ are the focal lengths of the depth camera. For this study, we use the parameters provided by~\cite{silberman2012indoor} without any re-calibration.

\begin{table}[t!]
\centering
\tabcolsep=0.09cm
\caption{List of handheld object classes and types of objects included in each class.}
 \begin{tabular}{| c | c | c |} 
 \hline
 Class ID & Class Label & Objects Included\\
 \hline\hline
 0 & no object & -\\ 
 1 & smartphone & cellphones, smartphones\\ 
 2 & tablet & iPad, Kindle, tablets\\ 
 3 & drink & cups, bottles, mugs, flasks\\ 
 4 & book & newspapers, books, sheets of paper\\ 
 \hline
 \end{tabular}
 \label{table:objects}
\end{table}

\subsection{Acquiring Handheld Object Labels}
In addition to locating hand instances in depth images, we also aim to identify objects in driver hands. To facilitate training for the same, we label each hand instance (obtained using Algorithm~\ref{alg:instances}) in a semi-automatic manner with minimal human input. We do this by following the procedure detailed in Algorithm~\ref{alg:objects}, the only requirement being that the driver (subject) holds the same object in the same hand for the entirety of a captured sequence. Multiple sequences are then captured with different subjects and different objects in each hand, one hand at a time. The variety of handheld objects considered for this study and the semantic classes they fall under are listed in Table~\ref{table:objects}.

\section{HandyNet}

\subsection{Network Architecture}

\begin{figure}[t]
\begin{center}
\includegraphics[width=0.7\linewidth]{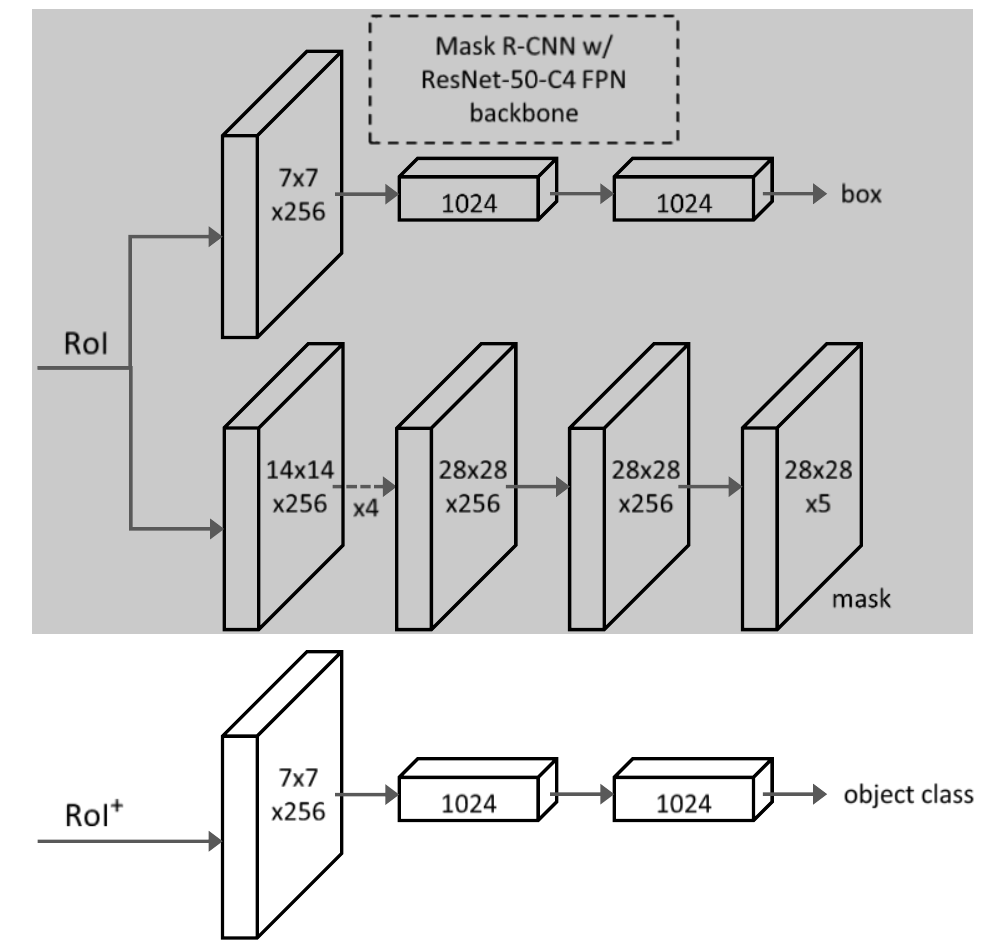}
\end{center}
\caption{\textbf{Head Architecture of HandyNet:} We retain the mask head from Mask R-CNN~\cite{he2017mask} but separate the classification and bounding box regression head. The classification head receives a larger region of interest (RoI$^+$) to better identify larger handheld objects.}
\label{fig:handynet}
\end{figure}

HandyNet is largely based on the state-of-the-art Mask R-CNN\cite{he2017mask} architecture. This architecture consists of two stages in sequence. First, the Region Proposal Network (RPN) generates class agnostic regions of interest (RoIs). We use the first 4 convolutional stages of ResNet-50 with features pyramids~\cite{lin2017feature} as the backbone for this purpose. The input to this network is the undistorted, registered, in-painted depth image as the sole channel. 

The second stage of the Mask R-CNN architecture is made of task specific heads. In our HandyNet architecture, we retain the structure of the mask head, and split the bounding box regression and classification head into two separate heads. While the mask and bounding box regression heads receive the same RoI from the RPN, the classification head receives a slightly larger region of interest (RoI$^+$). This follows from the reasoning that unlike conventional object classification, we are attempting to classify the handheld object for a given hand instance. This slightly larger RoI$^+$ might be favorable for identifying larger objects like tablets and newspapers. For an RoI parametrized by $(x, y, w, h)$, we define RoI$^+$ parametrized by $(x', y', w', h')$ as follows:

\begin{equation}
\begin{array}{l}
x'=x-\alpha \cdot w,\ y'=y-\alpha \cdot h,\\
\\
w'=(1+2\alpha) \cdot w,\ h'=(1+2\alpha) \cdot h,
\end{array}
\end{equation}

where $\alpha$ is the factor by which the region of interest is expanded. We choose the value of $\alpha$ based on cross validation.

\subsection{Implementation Details}

As in Mask R-CNN, an RoI is considered positive if it has IoU with a ground-truth box of at least $0.5$ and negative otherwise. We feed the input images at full-resolution i.e. without resizing. The training and inference configurations for HandyNet are listed in Table~\ref{table:handynet}. We make changes to the original configurations from Mask R-CNN to account for the average number of hand instances in a typical depth image, the scale of the hand instances encountered, and the more structured nature of the task at hand. All variants of the proposed network are trained for a total of $120$ epochs using the following schedule: First, the RPN (backbone) is trained for $40$ epochs with a learning rate of $0.001$. Next, the fourth stage of the ResNet-50 backbone (RPN) along with all task specific heads are trained for an additional $40$ epochs with the same learning rate. Finally, the entire network is trained for $40$ epochs with a reduced learning rate of $0.0001$. We use a momentum of $0.9$ and a weight decay of $0.0001$ throughout. Note that HandyNet is initialized from scratch with random weights. This is made possible by the huge labeled dataset available for training (see Table~\ref{table:dataset}).

\begin{table}[t!]
\centering
\tabcolsep=0.09cm
\caption{Training and inference configurations used\\for HandyNet.}
\begin{threeparttable}
 \begin{tabular}{| c | c |} 
 \hline
 \thead{Input image size} & $424\times512$\\
 \thead{Batch size} & $6$\\
 \thead{RPN$^1$ anchor scales} & $\{16, 32, 64, 128\}$\\
 \thead{RPN$^1$ anchor aspect ratios} & $\{0.5, 1, 2\}$\\
 \thead{Number of anchors per\\image used for RPN$^1$ training} & $64$\\
 \thead{Number of RoIs per image\\retained for training the heads} & $20$\\
 \thead{Ratio of positive RoIs\\ per image} & $0.1$\\
 \thead{RoIs retained post NMS\\ during training} & $100$\\
 \thead{RoIs retained post NMS\\ during inference} & $50$\\
 \hline
 \end{tabular}
 \begin{tablenotes}
\item[1] Regional Proposal Network
\end{tablenotes}
\end{threeparttable}
 \label{table:handynet}
\end{table}

\section{Experimental Analysis}

\begin{figure*}[t]
\begin{center}
\includegraphics[width=0.99\linewidth]{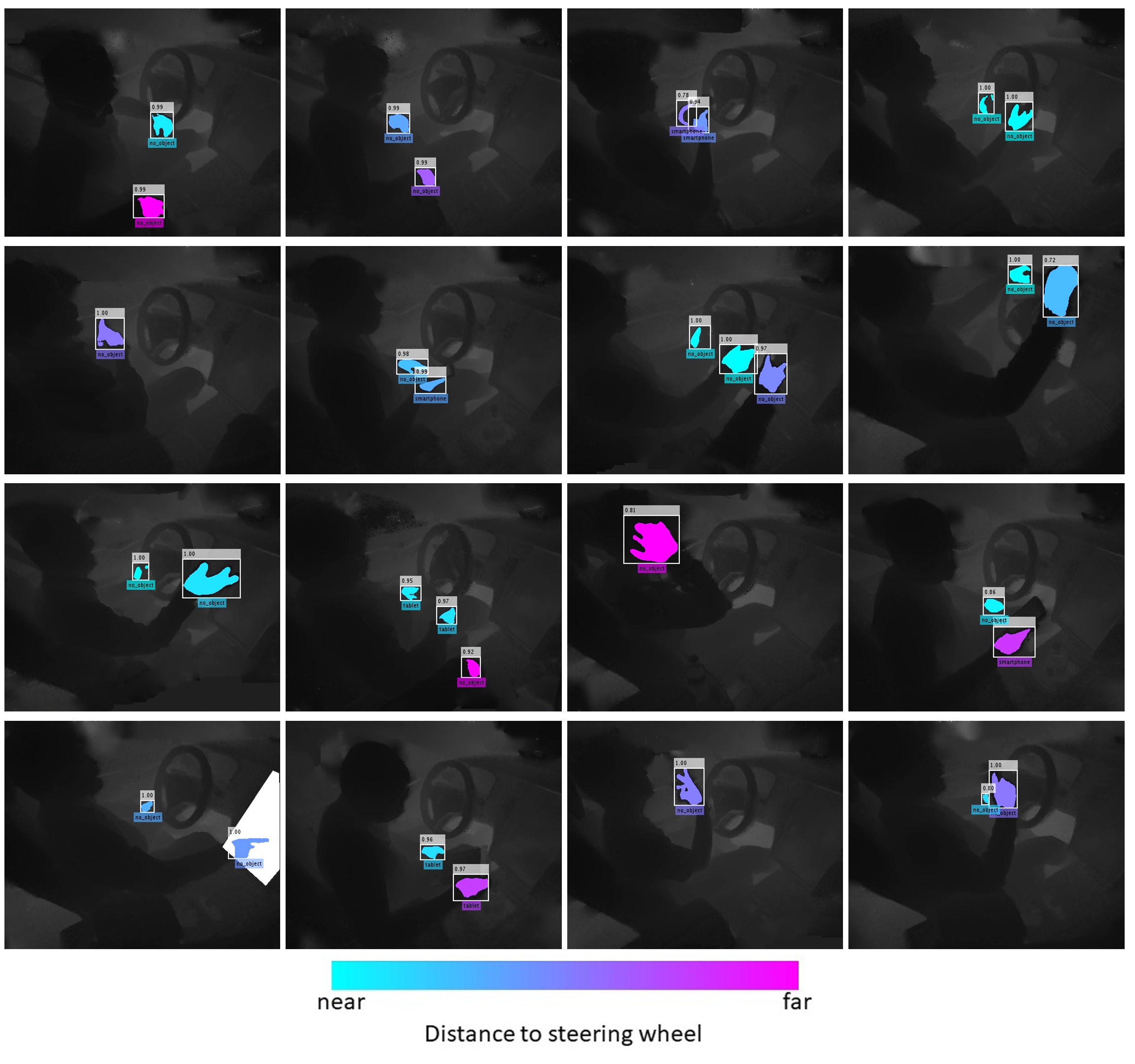}
\end{center}
\caption{\textbf{HandyNet results on naturalistic driving data (best viewed in color):} Results consist of drivers and objects not part of the training, validation and test sets. Operating on depth data allows us to gauge accurate distances to control elements like the steering wheel (see colorbar above) and the interactive display (see bottom left image). For our experiments, the steering wheel and interactive display were labeled by a human as part of a one-time calibration setup.}
\label{fig:results}
\end{figure*}

To test the viability of our semi-automatic labeling approach, we split the entire data as follows: the training and validation sets were mostly captured indoors with suitable lighting for chroma-keying. All subjects were wearing green gloves with red wrist bands as described in Section~\ref{sec:chromakey}. The subjects were asked to imitate naturalistic driving while holding different objects. The testing set is mostly comprised of data from real world drives i.e. the subject is actually driving the car. The subjects may interact with different objects as they would normally do in a drive. The remaining data in the testing set is captured indoors in a manner similar to the training/validation set. This is done to ensure that all objects included in the training set are covered in the test set as well. Moreover, all data is captured on a Tesla Model S testbed, where the drivers can take their hands off the wheel for extended periods of time. More details on each split are provided in Table~\ref{table:dataset}. In addition to this, we provide some qualitative results in Figure~\ref{fig:results} from completely naturalistic drives i.e. real drives where subjects are not wearing gloves or wrist bands. We also ensure that no drivers (subjects) overlap between different splits.


\begin{table}[t]
\centering
\tabcolsep=0.09cm
\caption{Details of the train-val-test split used for the experiments.}
 \begin{tabular}{| c | c | c | c | c |} 
 \hline
 \thead{Split} & \thead{Number of\\Unique Drivers} & \thead{Number of\\Frames} & \thead{Number of\\Hand Instances} & \thead{Fraction of\\Naturalistic\\Driving Data}\\
 \hline\hline
 \thead{Training} & $7$ & $128,317$ & $219,369$ & $0.2$\\
 \hline 
 \thead{Validation} & $1$ & $6,897$ & $13,525$ & $0.0$\\
 \hline 
 \thead{Testing} & $2$ & $36,497$ & $69,794$ & $0.9$\\
 \hline
 \end{tabular}
 \label{table:dataset}
\end{table}

\subsection{Timing}

\textbf{Training:} HandyNet with a ResNet-50-C4 RPN takes about 52 hours to train from scratch on our system with a 6 core Intel 990X processor and a Titan X Maxwell GPU. The training data is stored in a SATA III Solid State Drive (SSD).

\textbf{Inference:} Inference for HandyNet using the same system above runs at approximately $15$Hz. This includes the time for fetching and pre-processing the data. The relatively smaller RPN (with ResNet-50-C4 backbone), fewer number of anchor scales, and the fewer number of RoIs after non-maximal suppression all contribute to making the network run faster during inference.

\subsection{Ablation Experiments}

We perform comprehensive ablations on HandyNet using the validation and testing splits. We report standard COCO metrics~\cite{lin2014microsoft} including AP (averaged over IoU thresholds), AP$_{50}$, AP$_{75}$, and AP$_S$, AP$_M$ (AP at small and medium scales). We do not provide AP$_L$ (AP for large objects) due to the lack of large instances. Note that AP is evaluated using mask IoU. 

First, we determine the value of the hyperparameter $\alpha$ through cross validation. Table~\ref{table:result1} lists the various APs for $\alpha \in \{0.0, 0.1, 0.2, 0.3, 0.4, 0.5, 0.6\}$ on the validation split. We see that expanding the RoI improves the overall performance consistently until $\alpha=0.5$, after which we observe a saturation point. Increasing $\alpha$ beyond $0.5$ only seems to benefit object classification for very small hand instances, which are quite seldom. Hence, it was sensible for us to ignore such rare cases and optimize the performance for the more commonly observed hand sizes. With this reasoning, we chose HandyNet with $\alpha=0.5$ for reporting results on the testing set.


\begin{table}[t]
\centering
\tabcolsep=0.09cm
\caption{\textbf{Ablation results on validation split:} Mask AP for HandyNet with different values of expansion factor $\alpha$.}
 \begin{tabular}{ c | c c c | c c c } 
 \thead{$\alpha$} & \thead{AP} & \thead{AP$_{50}$} & \thead{AP$_{75}$} & \thead{AP$_S$} & \thead{AP$_M$}\\
 \hline
  $0.0$ & $28.8$ & $48.7$ & $26.0$ & $24.7$ & $43.3$\\
  $0.1$ & $28.9$ & $49.1$ & $26.5$ & $25.1$ & $43.4$\\
  $0.2$ & $29.2$ & $49.3$ & $27.1$ & $25.4$ & $43.8$\\
  $0.3$ & $29.9$ & $49.8$ & $27.3$ & $26.2$ & $43.9$\\
  $0.4$ & $30.0$ & $49.7$ & $27.7$ & $28.6$ & $44.0$\\
  $0.5$ & \textbf{30.6} & \textbf{51.8} & \textbf{28.0} & $29.4$ & \textbf{44.9}\\
  $0.6$ & $30.5$ & $51.7$ & $27.8$ & \textbf{29.6} & $44.5$\\
 \end{tabular}
 \label{table:result1}
\end{table}

Next, we provide both class agnostic and class sensitive results for the best performing model on the testing split (Table~\ref{table:result2}). Note that the class in our task is the associated handheld object class and not the semantic class itself. We see that the class agnostic performance is robust for all AP metrics, indicating our networks capability to successfully localize and segment driver hands. This also proves that our network generalizes well to naturalistic driving data captured with drivers not part of the original training split. As expected, the class sensitive APs are lower in comparison to the class agnostic ones, but not by much. Based on the qualitative and quantitative evaluation we have conducted, we believe that our network successfully identifies objects that are not too dissimilar to the objects it has been trained on. However, for images with considerably different handheld objects, or in situations where the objects are not completely visible due to the manner in which they are grasped, our network fails to produce the correct output. These issues could probably be solved by either gathering more diverse data, changing the camera view, or by incorporating temporal information. We leave these questions for future work.

\begin{table}[t]
\centering
\tabcolsep=0.09cm
\caption{\textbf{Class agnostic and class sensitive results on testing split:} Mask AP for best performing model ($\alpha=0.5$).}
 \begin{tabular}{ c | c c c | c c c } 
 \thead{Type of\\evaluation} & \thead{AP} & \thead{AP$_{50}$} & \thead{AP$_{75}$} & \thead{AP$_S$} & \thead{AP$_M$}\\
 \hline
 \thead{class agnostic} & $42.9$ & $83.3$ & $40.4$ & $34.7$ & $50.8$\\
 \thead{class sensitive} & $30.3$ & $51.2$ & $27.9$ & $28.5$ & $44.0$\\
 \end{tabular}
 \label{table:result2}
\end{table}


\section{Concluding Remarks}

In this study, we present HandyNet - a CNN that uses depth images to execute hand related tasks that may be of use in autonomous and semi-autonomous vehicles of the future. This includes detecting and localizing driver hands in 3D within a vehicle cabin, and additionally identifying handheld objects. Training such a network is made possible by our proposed method for semi-automatic labeling based on chroma-keying. The entire data used to train HandyNet from scratch ($128,317$ images and $219,369$ hand instances) was captured and labeled within a single day. This demonstrates the ease with which similar networks can be trained for new environments and different camera views. We hope this work inspires more ways to produce cheaply labeled data for related tasks, especially when the alternative is several hundred hours of human effort. 

\section{Acknowledgments}
We gratefully acknowledge all our sponsors, and especially Toyota CSRC for their continued support. We would also like to thank our colleagues at the Laboratory for Intelligent and Safe Automobiles (LISA), UC San Diego for their assistance in capturing the proposed dataset.

{\small
\bibliographystyle{ieee}
\bibliography{main_arxiv}
}

\end{document}